\documentclass{osa-article}

\journal{oe}
\articletype{Research Article}
{\begin{list}               %    with flush left bullets.
    {$\bullet$ \hfill}{
        \setlength{\leftmargin}{\parindent}
        \setlength{\parsep}{0.04\baselineskip}
        \setlength{\itemsep}{0.5\parsep}
        \setlength{\labelwidth}{\leftmargin}
        \setlength{\labelsep}{0em}}
    }
{\end{list}}

\providecommand{\eref}[1]{\eqref{#1}}  % call \eqref from amstex
\providecommand{\cref}[1]{Chapter~\ref{#1}}

\providecommand{\R}{\ensuremath{\mathbb{R}}}

\providecommand{\N}{\ensuremath{\mathbb{N}}}

\providecommand{\Pb}{\ensuremath{\mathbb{P}}}

\providecommand{\bydef}{\overset{\text{def}}{=}}

\renewcommand{\vec}[1]{\ensuremath{\boldsymbol{#1}}}
\providecommand{\mat}[1]{\ensuremath{\boldsymbol{#1}}}

\providecommand{\calD}{\mathcal{D}}

\providecommand{\calM}{\mathcal{M}}

\providecommand{\calT}{\mathcal{T}}

\providecommand{\mB}{\mat{B}}

\providecommand{\mG}{\mat{G}}

\providecommand{\mI}{\mat{I}}

\providecommand{\mS}{\mat{S}}

\providecommand{\mY}{\mat{Y}}
\providecommand{\mZ}{\mat{Z}}

\providecommand{\vb}{\vec{b}}
\providecommand{\vc}{\vec{c}}

\providecommand{\vu}{\vec{u}}
\providecommand{\vv}{\vec{v}}

\providecommand{\vx}{\vec{x}}
\providecommand{\vy}{\vec{y}}
\providecommand{\vz}{\vec{z}}

\providecommand{\vbeta}{\vec{\beta}}

\providecommand{\veta}{\vec{\eta}}
\providecommand{\vtheta}{\vec{\theta}}

\providecommand{\vztilde}{\boldsymbol{\widetilde{z}}}

\providecommand{\vhat}{\widehat{v}}

\newcommand{\argmin}[1]{\mathop{\underset{#1}{\mbox{argmin}}}}
\newcommand{\argmax}[1]{\mathop{\underset{#1}{\mbox{argmax}}}}

\begin{document}

\title{Megapixel photon-counting color imaging using quanta image sensor}

\author{Abhiram Gnanasambandam,\authormark{1,*} Omar Elgendy,\authormark{1} Jiaju Ma,\authormark{2} and Stanley H. Chan\authormark{1}}

\address{\authormark{1}School of Electrical and Computer Engineering, Purdue University, West Lafayette, IN 47907, USA\\
\authormark{2} Gigajot Technology Inc., Pasadena, CA 91107, USA}

\email{\authormark{*}  agnanasa@purdue.edu} %% email address is required

\homepage{https://engineering.purdue.edu/ChanGroup/} %% author's URL, if desired


\begin{abstract}
Quanta Image Sensor (QIS) is a single-photon detector designed for extremely low light imaging conditions. Majority of the existing QIS prototypes are monochrome based on single-photon avalanche diodes (SPAD). Passive color imaging has not been demonstrated with single-photon detectors due to the intrinsic difficulty of shrinking the pixel size and increasing the spatial resolution while maintaining acceptable intra-pixel cross-talk. In this paper, we present image reconstruction of the first color QIS with a resolution of $1024 \times 1024$ pixels, supporting both single-bit and multi-bit photon counting capability. Our color image reconstruction is enabled by a customized joint demosaicing-denoising algorithm, leveraging truncated Poisson statistics and variance stabilizing transforms. Experimental results of the new sensor and algorithm demonstrate superior color imaging performance for very low-light conditions with a mean exposure of as low as a few photons per pixel in both real and simulated images. 
\end{abstract}

%%%%%%%%%%%%%%%%%%%%%%%%%%  body  %%%%%%%%%%%%%%%%%%%%%%%%%%
\section{Introduction}
Quanta image sensor (QIS) is a concept proposed by E. Fossum in 2005 aiming to overcome the limited sensor performance with the continuously shrinking pixel size~\cite{fossum200611,fossum2005gigapixel}. In a CMOS image sensor \cite{fossum1993active}, images are acquired by accumulating hundreds to thousands of photoelectrons in each pixel and converting the charges to voltages and then to digital signals. As pixel size shrinks, the amount of photons that fall on individual pixels drops. Under low-light (or high-speed imaging) conditions, photon shot noise dominates the acquisition process. As the photon becomes scarce, eventually the signal-to-noise ratio will drop to a point that no meaningful images can be generated. Small pixel also reduces the full-well capacity which limits the dynamic range. QIS overcomes the problem by spatial-temporally oversampling the scene. In a QIS, each ``image pixel'' is partitioned into a group of smaller pixels called jots, and each jot is sensitive enough for photon counting. Over the past decade, numerous studies have demonstrated the feasibility of QIS, with a few prototypes developed \cite{ma2015pump,ma2017photon,hynecek2001impactron,dutton2014320,bruschini2018monolithic}. On the image reconstruction side, various methods have also been proposed \cite{Chan14,Chan16,Elg18,yang2012bits}.

Despite the advances in QIS hardware and image reconstruction algorithms, one important open problem is the color imaging capability of QIS. Color imaging in low light is important for many computer vision applications such as night-vision. It is also important in biological studies, where using brighter light may lead to degradation of the biological matter.  While active color imaging have been demonstrated using SPAD in \cite{Ren:18} passive color imaging using QIS is difficult for two reasons. First, the resolution of a typical single-photon detector array is not high, e.g., $320 \times 240$ pixels~\cite{dutton2016single}, or most recently $512 \times 512$ pixels ~ \cite{bruschini2018monolithic}. If a color filter array such as the standard Bayer pattern is implemented, the effective resolution will be halved to $160\times 120$ or $256\times256$. For detection/tracking applications \cite{gyongy2018single} this might be sufficient; but for photography the resolution could be low compared to the mainstream CMOS image sensors. Second, the color reconstruction of QIS data is different from CMOS image sensors because the QIS data is acquired under sparse-photon conditions with much lower signal-to-noise ratio. The demosaicing algorithm of QIS data needs to overcome the photon shot noise and accurately reproduce the color information. As we will discuss, QIS requires a joint reconstruction and demosaicing on top of the truncated Poisson statistics, a problem that has not been studied.

\begin{figure}[t]
\centering
\begin{tabular}{cccc}
\includegraphics[width=0.23\linewidth]{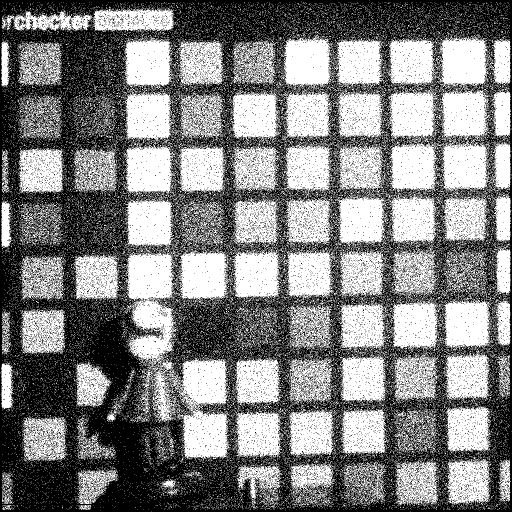}&
\hspace{-2.0ex}
\includegraphics[width=0.23\linewidth]{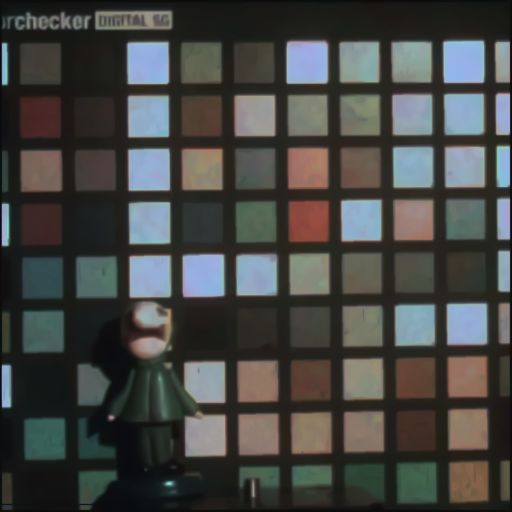}&
\hspace{-2.0ex}
\includegraphics[width=0.23\linewidth]{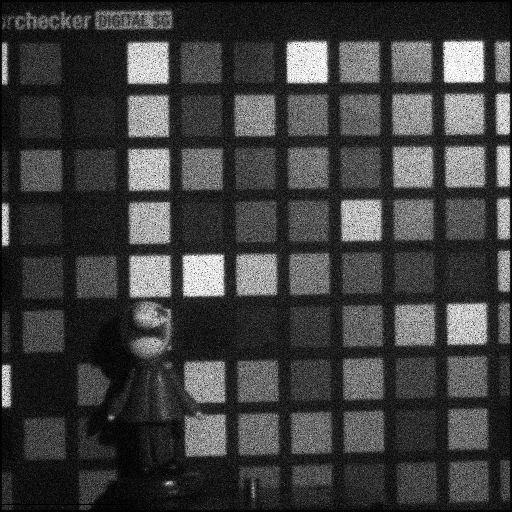}&
\hspace{-2.0ex}\includegraphics[width=0.23\linewidth]{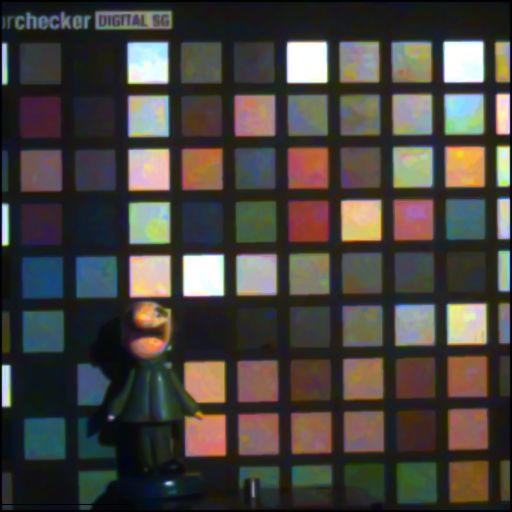}\\
\footnotesize  (a) Raw input, 1-bit &
\hspace{-2.0ex}
\footnotesize (b) Processed from (a) &
\hspace{-2.0ex}\footnotesize (c) Raw input, 5-bit&
\hspace{-2.0ex}\footnotesize (d) Processed from (c) \\
\end{tabular}
\caption{(a) One 1-bit frame. (b) Reconstructed color image using \textcolor{black}{50 frames} of 1-bit input with threshold $q=4$. (c) One 5-bit frame. (d) Reconstructed color image using \textcolor{black}{10 frames} of 5-bit input. The average number of photons per frame is 5.}
\label{fig:recSingleMulti}
\vspace{-2.0ex}
\end{figure}

\begin{figure}[!t]
\centering
    \begin{tabular}{cc}
    \includegraphics[width=0.35\linewidth]{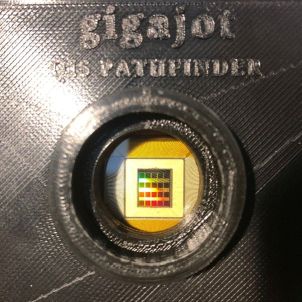}&
    \includegraphics[width=0.35\linewidth]{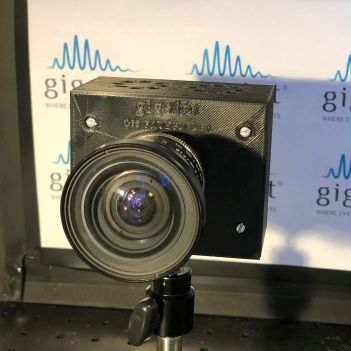}\\
    (a) QIS sensor chip & (b) Prototype QIS camera module
    \end{tabular}
    \caption{Gigajot prototype QIS camera module and the QIS sensor chip. Note that there is no additional optics required besides a simple focusing lens. The camera is powered by standard 5V DC input, and has a USB3 data interface to transmit data to external storage.}
\label{fig: camera module and sensor}
\end{figure}

In this paper, we report the first demonstration of QIS color imaging. The QIS Pathfinder camera module developed at Gigajot Technology (as shown in Fig. \ref{fig: camera module and sensor}) was used in this demonstration. The camera sensor has the same core design as the sensor described in \cite{ma2017photon}, but has a Bayer pattern color filter array implemented on-chip. It has a spatial resolution of $1024\times1024$, single-bit and multi-bit photon-counting outputs, and up to $1040$ fps speed in single-bit mode. Figure \ref{fig:recSingleMulti} shows a snapshot of the work presented in this paper. The key innovation, besides the sensor, is a new customized image reconstruction algorithm which enables joint reconstruction and demosaicing of raw QIS data. Our method leverages a few recently developed techniques in the QIS literature, including the transform-denoise method \cite{Chan16} and the Plug-and-Play  Alternating Direction Method of Multipliers (ADMM) \cite{chan2017plug}. We demonstrate both our new sensor and the new algorithm on real data, and make quantitative evaluation based on synthetic data.

\section{Megapixel quanta image sensor}\label{sec:hardware}
When the concept of QIS was first proposed, the two mainstream design methodologies to realize single-photon detection were the electron-multiplying charge-coupled device (EMCCD) \cite{hynecek2001impactron} and the single-photon avalanche diode (SPAD) \cite{aull2002geiger,buchholz2018widefield}. Both devices amplify the signal from a single photoelectron with a physical phenomenon called electron \emph{avalanche multiplication}, where high electrical voltage (typically higher than 20V) is used to accelerate the photoelectron to produce more free electrons through collision. SPAD is more widely adopted in consumer applications because of its excellent low-noise performance and temporal resolution. The SwissSPAD  \cite{charbon2008towards,charbon2007will,bruschini2018monolithic} and the SPAD developed at University of Edinburgh \cite{dutton2014320,dutton2016single} are two better known examples. However due to its fabrication technology SPAD often has to make compromise in at least one of the three factors - low dark current, high quantum efficiency, and small pixel pitch. The QIS we use in this paper is based on a new type of single-photon detector proposed by Ma \emph{et al.} \cite{ma2015pump,ma2015quanta,fossum2005gigapixel,ma2017photon,masoodian20171mjot}. The approach is to enhance the voltage signal generated by a single photoelectron by reducing the capacitance of the pixel output node so that the single-photon signal can overcome the background thermal noise. 
%However, due to the relatively large pixel size (typically more than 5$\mu$m pitch), high dark count, and low quantum efficiency, SPADs are not ideal candidates for implementing the QIS devices.

\begin{figure*}[t]
\centering
\includegraphics[width=\linewidth]{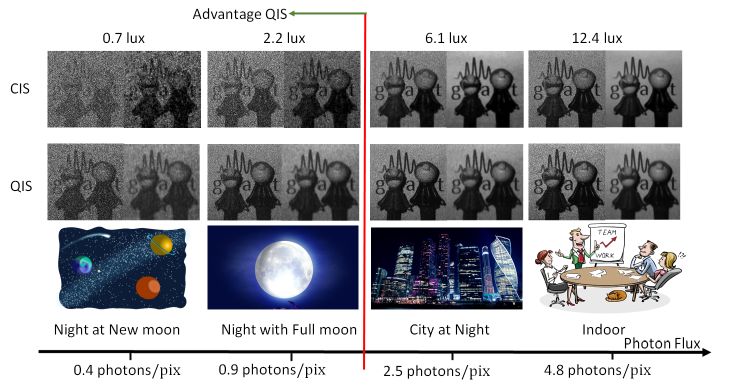}
\caption{Quanta Image Sensor (QIS) vs. CMOS Image Sensor (CIS). The top row shows a simulated CIS data at a photon level same as the QIS. The second row shows the real analog sensor data obtained from our prototype camera. For each photon flux level, we show both the raw input data and the denoised data. The CIS is assumed to have a read noise of $1.2 e^{-}$ r.m.s. , and we do not considered dark count. With dark count, the performance of CIS will deteriorate even more. }
\label{fig:IPhonevsQIS}
\end{figure*}

Table \ref{tab:Comparison} shows a comparison between a conventional CMOS image sensor, SPAD, and the prototype QIS used in this work. Among the listed different sensor parameters, the most noticeable difference between the QIS and SPADs is the small pixel pitch, where the QIS realized 1.1 $\mu$m pixel size, but SPADs typically have a pitch size around 5-20 $\mu$m. The smaller pixel size makes high spatial resolution (e.g. 10s or 100s of megapixel) possible with a small optical format (e.g. 1/4''). Another advantage of QIS over SPADs is the high quantum efficiency. Quantum efficiency measures the percentage of photons that are detectable by the sensor, which is critical for low-light imaging. 
SPAD cameras generally have a low quantum efficiency (e.g. $<$50\%), because of low fill-factor and low photon detection capability normally called photon detection efficiency aka PDE. While low fill factor can be alleviated to an extent by using optical solutions such as microlens light-pipe. This reduces the photosensitive areal percentage in the SPAD pixels and limits the quantum efficiency. In contrast, since electron avalanche gain is not required in the QIS device, it has a quantum efficiency over 80\% within the visible wavelengths. 

Figure \ref{fig:IPhonevsQIS} shows comparison between real QIS image and a simulted CIS image. We note that we assume that the CIS no dark current. Even without considering the advantages of a lower dark current, the performance of QIS in photon starved regime is much better than the CIS.

\begin{table*}
\footnotesize
\caption{\label{tab:Comparison}Comparison of the available image sensor technologies.}
\makebox[1 \textwidth][c]{       %centering table
  \begin{tabular}[b]{ c  c  c c c c}
    \textbf{Camera} & \textbf{ Sony IMX253} & \textbf{ Edin. SPAD}  & \textbf{SwissSPAD2 } & \textbf{QIS}  & \textbf{sCMOS}     \\
    &CMOS&\cite{dutton2016spad}&\cite{bruschini2018monolithic}&(This work)&\cite{scmos}\\
    \hline \hline
    \textbf{Resolution} & 4096 $\times$ 3000 &320 $\times$ 240 &   512 $\times$ 512& 1024 $\times$ 1024 & 2048 $\times$ 2048 \\
    \textbf{Mean Dark Count Rate} &  2.4 $e^{-}$/pix/s& 47 $e^{-}$/pix/s &  75 $e^{-}$/pix/s & 0.068 $e^{-}$/pix/s & 0.1 $e^{-}$/pix/s at $0^\circ C$\\
    \textbf{Quantum Efficiency} & $72\%$ & $50\%$    & $39.5\%$    & $86\%$   & $82\%$  \\
    & at 525 nm & at 520 nm & at 480 nm & at 480 nm & at 560 nm\\
    \textbf{Pixel Pitch} &3.45$\mu$m & 8 $\mu$m& 16.4 $\mu$m & 1.1 $\mu$m & $6.5\mu m$ \\
    \textbf{Frames / sec}  & 30 & $16$k & 97.7k &1040 & 30\\
    \textbf{Photon Counting}  &No & Yes &Yes &Yes & No\\
    \textbf{Color Capability} &Yes &No &No &Yes & Yes\\
    \hline
  \end{tabular}
  }
[Note : Measurements under room temperature unless otherwise noted]

\end{table*}

\section{QIS color imaging model}
\label{sec:imagingModel}
\begin{figure*}[!]
    \centering
    \includegraphics[width=\linewidth]{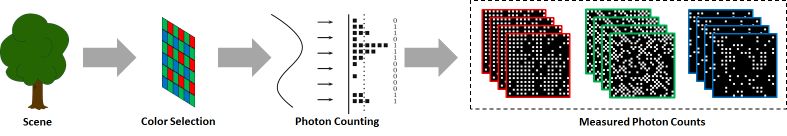}
    \caption{QIS Imaging Model. When the scene image arrives at the sensor, the color filter array first  selects the wavelength according to the colors. Each color pixel is then sensed using a photon-detector. In single-bit mode the detector reports a binary value, and in multi-bit mode the detector reports an integer up to the saturation limit. The measured data contains three subsampled sequences, each representing a measurement in the color channel.}
    \label{fig:Imgaing_model}
\end{figure*}

The imaging model of a color QIS is illustrated in Fig. \ref{fig:Imgaing_model}. Relative to the previous QIS imaging models, e.g., \cite{yang2012bits,Chan16,Elg18}, the new model considers a color filter array which allows us to select  wavelengths and multiple color channels. There are a few essential components in the model.

\vspace{1ex}
\noindent\textbf{Spatial oversampling}. We represent the discretized scene image as a column vector $\vc = [\vc_r; \vc_g; \vc_b] \in [0,1]^{3N}$, where $N$ is the number of pixels, $\vc_r$, $\vc_g$, and $\vc_b$ are the light intensities of the red, green and blue channels, respectively. QIS was originally designed as a spatial-temporally oversampling device which samples each pixel using $K$ jots spatially. This means every pixel has $K$ measurements, and totally there are $M \bydef NK$ measurements. Thus, the model of this oversampling process is
\begin{equation}
\vv = [\vv_r; \vv_g; \vv_b] = \alpha \mG\vc \in \R^{3M},
\end{equation}
where $\mG \in \R^{3M \times 3N}$ represents the $K$-fold upsampling and interpolation filtering. The constant $\alpha$ is a gain factor. In our current prototype, the oversampling factor is $K=1$ as we want to maximize the spatial resolution. In this case, we have $\mG = \mI$, and $M=N$.

\vspace{1ex}
\noindent\textbf{Color filter array}. To obtain color information, a color filter is placed on the top of each jot to collect light for only one of the RGB colors, and filter out the remaining two colors. This process is represented by three mutually exclusive $M\times M$ diagonal matrices or masks: $\mS_r$, $\mS_g$ and $\mS_B$ for the red, green and blue channels, respectively. These matrices contain either zeros or ones on the diagonal, and $\mS_r + \mS_g + \mS_b = \mI$.

Incorporating the color filter array, the light exposure on the $M$ jots of QIS is represented by
\begin{equation}\label{eq:model}
\vtheta = \mS_r\vv_r+\mS_g\vv_g+\mS_b\vv_b \;\;\in \mathbb{R}^{M}.
\end{equation}
We call $\vtheta$ the mosaiced image. If we let $\mS \bydef [\mS_r, \; \mS_g, \; \mS_b] \in \R^{M \times 3M}$, then $\vtheta$ and $\vc$ are related by
\begin{equation}
    \vtheta = \alpha\mS\mG\vc.
\end{equation}

\vspace{1ex}
\noindent\textbf{Photon arrival}. The photon arrival of QIS is modeled as a Poisson process. Let $\mY \in \N^{M}$ be a non-negative random integer denoting the number of photons arrived at each jot during one integration period. Then, the probability of observing $\mY = \vy \in \N^{M}$ is
\begin{equation}\label{eq:prob_Ym}
\Pb(\mY = \vy)= \prod_{m=1}^{M} \frac{\theta_m ^{y_m} e^{-\theta_m}}{y_m!}.
\end{equation}
For single-bit QIS, we put a voltage comparator to threshold the arrived photon count into a binary decision. Thus, the read out of each jot is a binary random variable $\mB \in \{0,1\}^{M}$ with $B_m = 1$ if $Y_m \ge q$ and $B_m = 0$ if $Y_m < q$, where $q > 0$ is a threshold.

For multi-bit QIS, we use an
$L$-bit quantizer to quantize the measurement. In this case,
\begin{equation}
  B_m =
    \begin{cases}
    Y_m,& Y_{m} < 2^{L}-1, \\
     2^{L}-1,& Y_{m} \geq  2^{L}-1.
    \end{cases}
\end{equation}
In other words, we count the photons until the number reaches a threshold. When the threshold is reached, we report the maximum number before saturation.

\begin{figure}[!t]
\centering
\includegraphics[width=0.6\linewidth]{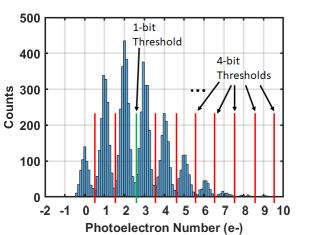}
\caption{Measured photo-electron count from a real sensor. The non-integer values of the count are due  to the read noise of the readout circuit. However, with the ultra-low read noise provided by the QIS device, its negative effect on the photon-counting is negligible by setting appropriate thresholds.}
\label{fig:photon count histogram}
\end{figure}

\vspace{2ex}
\noindent\textbf{Assumptions}. While the above model is theoretically tractable, we made three assumptions. The first is regarding the read noise associated with $\mY$. If read noise is present, the measured photon count will become $\vy + \veta$ where $\veta$ is typically i.i.d. Gaussian. We assume that the read noise is negligible compared to the photon count, and can be eliminated by the thresholding process. We find that this is often the case with the QIS device used in this paper, which provides ultra-low read noise (0.24e- rms at room temperature) and accurate photon-counting capability. See Fig. \ref{fig:photon count histogram} for an actual measurement of the photon count.

The second assumption we make is that the multi-bit saturation does not happen. That is, during the multi-bit sensing, we will never reach the upper threshold $q$ and so $B_m = Y_m$, which is a standard Poisson random variable. In terms of hardware, this can be achieved by implementing an automatic exposure control scheme so that the threshold $q$ is increased when the scene is too bright. We will leave this to a follow up paper as it is outside the scope of this work.

The third assumption we make is that the camera is fast enough to catch up the scene movement, and the scene remains static during the multiple temporal samplings. This allows us to utilize multiple independent measurements over time to improve the statistics. In this case, the probability distribution of $\mY$ becomes
\begin{equation}\label{eq:prob_Ym 2}
\Pb(\mY = \vy)= \prod_{t=1}^T \prod_{m=1}^{M} \frac{\theta_m ^{y_{m,t}} e^{-\theta_m}}{y_{m,t}!},
\end{equation}
where $y_{m,t}$ is the Poisson count at pixel $m$ in time $t$.

\begin{figure*}[!]
    \centering
    \includegraphics[width = 1\linewidth]{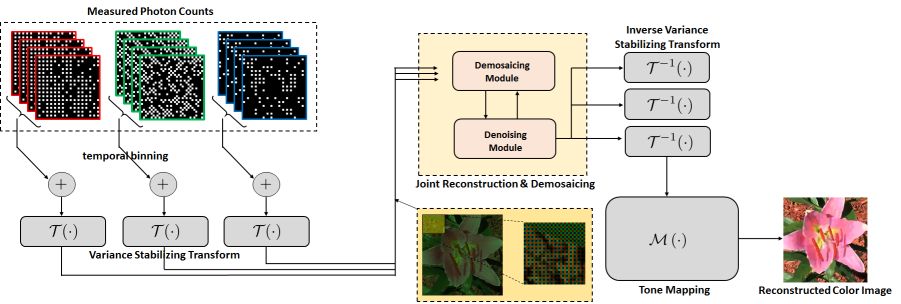}
    \caption{The image reconstruction pipeline consists of (i) a temporal binning step to sum the input frames, (ii) a variance stabilizing transform $\calT$ to transform the measurement so that the variance is stabilized, (iii) a joint reconstruction and demosaicing algorithm to recovery the color, (iv) an inverse transform to compensate the forward transform, and (v) a tone mapping operation to correct the contrast.}
    \label{fig:color_image_pipeline}
\end{figure*}

\section{QIS color image reconstruction} \label{sec:pipeline}
The task of image reconstruction is to recover color scene $\vc$ from the measurements $\mB$. In the gray-scale setting, we can formulate the problem as maximum-likelihood and solve it using convex optimization tools  \cite{yang2010optimal,yang2012bits,Chan14,Elg18}. We can also use learning-based methods, e.g., \cite{Rem16,choi2018image,chandramouli2018alittle,rojas2017learning} to reconstruct the signal. The method we present here is based on the transform-denoise approach by Chan \emph{et al.} \cite{Chan16}. We prefer transform-denoise because it is a physics-based approach and is robust to different sensor configurations. For example, in learning-based approaches, if we change the number of frames to sum, we need to train a different model or neural network.

\subsection{Reconstruction pipeline}
The pipeline of the proposed reconstruction algorithm is shown in Fig. \ref{fig:color_image_pipeline}. Given the random variable $\mB = \{B_{m,t}\}$ with realizations $\vb = \{b_{m,t}\}$, we first sum the frames to generate a single image $\mZ = \{Z_{m}\}$:
\begin{equation}
    Z_m = \sum_{t=1}^T B_{m,t}.
\end{equation}
In the future, this step can be integrated into the hardware of the camera, so that the output of the camera will be the sum of multiple frames.
Depending on whether we are using the single-bit mode or the multi-bit mode, the statistics of $\{Z_m\}$ follows either a Binomial distribution or a Poisson distribution. In single-bit mode, the distribution of $\{Z_m\}$ is
\begin{equation}
    \Pb[Z_m = k \;|\; \theta_m] = {T \choose k} \Psi_q(\theta_m)^{1-k}(1-\Psi_q(\theta_m))^{k},
\end{equation}
where $\Psi_q(\theta) \bydef \Pb[B = 0] = \sum_{j=0}^{q-1} \theta^j e^{-\theta}/j!$ is the probability of obtaining a zero in the binary variable $B$. As shown in \cite{Elg18}$, \Psi_q$ is the incomplete Gamma function. In multi-bit mode, since a sum of Poisson remains a Poisson, the distribution of $\{Z_m\}$ is
\begin{equation}
    \Pb[Z_m = k \;|\; \theta_m] = \frac{(T\theta_m)^k e^{-T\theta_m}}{k!}.
\end{equation}

If we assume that there is no processing of the bits besides temporal summation, then a simple maximum-likelihood of the probability will return us the solution:
\begin{align}
    \vtheta &= \argmax{\vtheta}\;\; \prod_{m=1}^M \Pb[Z_m = z_m \;|\; \theta_m] \notag \\
    &\bydef \calM(\vec{z}) =  \begin{cases}
    \Psi_q^{-1}(1-\vec{z}/T), &\quad \mbox{Single-bit},\\
    \vec{z}/T, &\quad \mbox{Multi-bit}.
    \end{cases}
\end{align}
The operation $\calM$ can be interpreted as a tone-mapping.

%As a special, if $q=1$ then $\Psi_q^{-1}(z) = \log(1-z)$.

\subsection{Variance stabilizing transform}
Now, we want to process the data besides just summing the frames. However, in either single-bit or multi-bit mode, the Binomial or the Poisson statistics are not easy as the variance changes with the mean. This prohibits the use any off-the-shelf algorithms that are based on i.i.d. Gaussian assumptions, e.g., denoising or inpainting.

The variance stabilizing transform (VST) is a statistical technique that tries to alleviate the changing variance problem \cite{azzari2017variance, azzari2016variance,foi2011noise}. Among the different types of VSTs, we are particularly interested in the two transforms originally proposed by F. Anscombe in 1948 \cite{Anscombe_1948}:
\begin{equation*}
\vbeta = \calT(\vz) \bydef
\begin{cases}
\sqrt{T+\frac{1}{2}}\sin^{-1}\sqrt{\frac{\vz+\frac{3}{8}}{T + \frac{3}{4}}}, &\quad\mbox{Single-bit},\\
\sqrt{\vz + \frac{3}{8}}, &\quad\mbox{Multi-bit}.
\end{cases}
\end{equation*}
Here, the first transform is called the Anscombe Binomial, and the second is called the Anscombe Poisson. Both transforms are pixel-wise, meaning that we can apply the transform to each pixel independently. As proved in \cite{Chan16} and \cite{Anscombe_1948}, the variance of $\calT(\vz)$ is 1/4 for both the single-bit and the multi-bit case. The inverse $\calT^{-1}$ is simply the algebraic inverse (or the unbiased inverse if we follow \cite{Makitalo_Foi_2011}).

The important question for color imaging is now that the photon count $\vz$ is the result of applying a color filter array matrix $\mS = [\mS_r,\mS_g,\mS_b]$. That is, instead of observing the full color $\vztilde \bydef [\vz_r;\vz_g;\vz_b]$, we only observe a subsampled version $\vz = \mS\vztilde$. However, since $\calT$ is point-wise and $\mS$ is a concatenation of three non-overlapping binary matrices, it holds that
\begin{equation}
    \calT(\vz) = \calT(\mS\vztilde) = \mS\calT(\vztilde).
\end{equation}
Consequently, if we define $\calT(\vz)$ as the observed measurement (which is available from the sensor) and treat $\vx \bydef \calT(\vztilde)$ as the unknown quantity to be determined, then the problem can be formulated as
\begin{equation}
    \vhat = \argmin{\vx}\;\; \|\mS\vx - \calT(\vz)\|^2 + \lambda g(\vx),
    \label{eq: demosaic-denoise}
\end{equation}
where $g$ is some regularization function controlling the smoothness of $\vx$. Note that \eref{eq: demosaic-denoise} is a standard demosaicing-denoising problem assuming i.i.d. Gaussian noise.

\subsection{Joint reconstruction and demosaicing}
The optimization problem in \eref{eq: demosaic-denoise} is a standard least squares with regularization function $g$. Thus, most convex optimization algorithm can be used as long as $g$ is convex. In this paper, we adopt a variation of the alternating direction method of multiplier (ADMM) by replacing $g$ with an off-the-shelf image denoiser. Such algorithm is coined the name Plug-and-Play (PnP) \cite{venkatakrishnan2013plug} (and different versions  thereafter \cite{chan2017plug}).

For the particular problem in \eref{eq: demosaic-denoise}, the PnP ADMM algorithm iteratively updates the following two steps:

\vspace{1ex}
\noindent\textbf{Demosaicing Module}:
\begin{equation}
    \vx^{(k+1)} = (\mS^T\mS + \rho\mI)^{-1}(\mS^T\calT(\vz) + \rho(\vv^{(k)}-\vu^{(k)})),
\end{equation}
\noindent\textbf{Denoising Module}:
\begin{equation}
    \vv^{(k+1)} = \calD_{\rho/\lambda}(\vx^{(k+1)}+\vu^{(k)}),
\end{equation}
and updates the Lagrange multiplier by $\vu^{(k+1)} = \vu^{(k)} - (\vx^{(k+1)}-\vv^{(k+1)})$. Readers interested in the detailed derivation can consult, e.g., \cite{chan2017plug}. Here, $\rho$ is an internal parameter that controls the convergence. The operator $\calD$ is an off-the-shelf image denoiser, e.g., Block-matching and 3D filtering  (BM3D) or deep neural network denoisers. The subscript $\rho/\lambda$ denotes the denoising strength, i.e., the hypothesized ``noise variance''. Since $\mS^T\mS$ is a diagonal matrix, the inversion is pointwise.

\subsection{Non-iterative algorithm}
\begin{figure}[h]
\centering
\begin{tabular}{cc}
\includegraphics[width=0.35\linewidth]{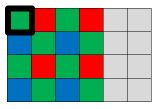}&
\includegraphics[width=0.35\linewidth]{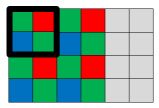}
\end{tabular}
\caption{In the future when QIS can achieve higher spatial resolution, we can use four color jots to reconstruct one pixel. In this case, we can bypass the iterative ADMM algorithm and use a one-shot denoising method.}
\label{fig:2x and 1x}
\end{figure}

The $1.1\mu$m pixel pitch of the proposed QIS can potentially lead to a spatial resolution as high as or even higher than a conventional CMOS sensor. When this happens, in certain applications we can trade-off the color reconstruction efficiency and the resolution. For example, instead of using one jot for one pixel, we can use four jots for one pixel as shown in Fig. \ref{fig:2x and 1x}.

Using four jots for one pixel allows us to bypass the iterative ADMM steps because there is no more missing pixel problem. In this case, the matrix $\mS \in \R^{M \times 3M}$ will become $\mS = \mbox{diag}\{\frac{1}{4}\mI, \frac{1}{2}\mI, \frac{1}{4}\mI\} \in \R^{3M \times 3M}$, and hence \eref{eq: demosaic-denoise} is simplified to a denoising problem with different noise levels for the three channels. In particular, the green channel has half of the variance of the red and the blue. For implementation, we can modify a denoiser, e.g., BM3D to accommodate the different noise variances. Since there is no more ADMM iteration, the algorithm is significantly faster. While we have used BM3D to demonstrate the results, any off the shelf denoiser which is used in CIS based cameras can be used to denoise the image for the four-jot to one-pixel method. We would also like to stress on the fact that both the Anscombe transform and the transform $\calM$ can be implemented as a look up table. So, this method can be as fast a current denoiser being used in a CIS based camera.

\subsection{Comparisons}
\begin{figure}[t]
\centering
\begin{tabular}{ccc}
\includegraphics[width=0.25\linewidth]{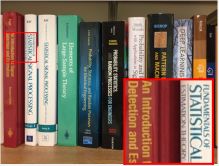} &
\includegraphics[width=0.25\linewidth]{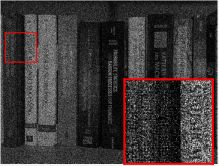}&
\includegraphics[width=0.25\linewidth]{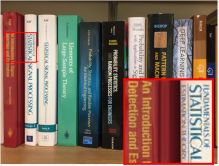} \\
(a) Ground truth & (b) Simulated Input  & (c) MATLAB $30.91$dB\\
\includegraphics[width=0.25\linewidth]{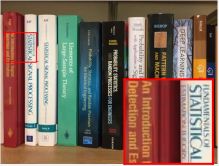} &
\includegraphics[width=0.25\linewidth]{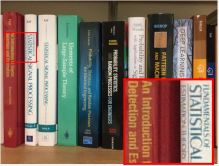} &
\includegraphics[width=0.25\linewidth]{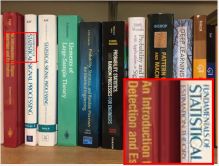}\\
(d) \cite{Jeon_Dubois_2013} $30.24$dB & (e) \cite{Hirakawa_Meng_Wolfe_2007} $29.57$dB & (f) Ours $31.51$dB
\end{tabular}
\vspace{0.5ex}
\caption{Simulated QIS experiment. The goal of this experiment is to compare the proposed iterative algorithm with existing methods. We assume the observed Bayer RGB image is from a 3-bit QIS sensor. (a) Ground Truth; (b) One 3-bit QIS frame; (c) MATLAB demosaic preceded and followed by BM3D; (d) LSLCD\cite{Jeon_Dubois_2013};  (e) Hirakawa's PSDD method \cite{Hirakawa_Meng_Wolfe_2007}, with a built-in wavelet shrinkage denoiser;  (f) Proposed method with BM3D.}
\vspace{-2ex}
\label{fig:demosaicing CMOS}
\end{figure}

We compare the proposed method with several existing methods on a synthetic dataset shown in Fig. \ref{fig:demosaicing CMOS}. We simulate the raw color QIS data by assuming a Bayer pattern and using the image formation pipeline described in the previous section. We demosaic the images using: (a) a baseline method using MATLAB's demosaic preceded by gray-scale BM3D denoising of $R$, $G_1$, $G_2$ and $B$ channels and followed by color BM3D denoising; (b) Least-squares luma-chroma demultiplexing (LSLCD) method \cite{Jeon_Dubois_2013}, which has a built-in BM3D denoiser; (c) Hirakawa's PSDD method \cite{Korneliussen_Korneliussen_2014,Hirakawa_Meng_Wolfe_2007}, which does joint denoising and demosaicing for Poisson noise; and (d) the proposed method using BM3D with $(\lambda,\rho)=(0.001,5)$. We apply variance stabilizing transform, except for PSDD which is designed for Poisson noise. The results show that the proposed method has a significantly better performance, both in terms of Peak Signal to Noise Ratio (PSNR) and visual quality.

%books_n_3_ADMM_rho_5_lambda_0.0010_BM3D_10iter_31.51dB_0.9790.jpg

\section{Experimental results}
\subsection{Synthesized QIS data}
We conduct a synthetic experiment to provide a quantitative evaluation of the performance of the proposed algorithm. To this end, we simulate the image formation pipeline by passing through color images to generate the QIS raw input data, with different number of bits. Figure \ref{fig:recSingleMultiSynth} shows one example, and in the Supplementary Report we have additional examples.

In our simulation, we assume that the number of QIS frames is $T = 4$, and the average number of photon per pixel is $0.28$, $0.85$, $1.98$ and $4.23$ photons / frame for 1-bit, 2-bit, 3-bit and 4-bit QIS, respectively. On the measurement side, we generate single-bit and multi-bit data by thresholding the raw sensor output. To reconstruct the image, we use the proposed method with PnP and BM3D. The parameters are set as $\rho=1$ and $\lambda=0.007, 0.003, 0.002$ and $0.0007$ for 1-bit, 2-bit, 3-bit and 4-bit QIS, respectively. With as low as 1-bit, the reconstructed image in Fig. \ref{fig:recSingleMultiSynth} is already capturing most of the features. As the number of bits increases, the visual quality improves.

\begin{figure*}[!t]
\centering
\begin{tabular}{cccc}
\includegraphics[width=0.23\linewidth]{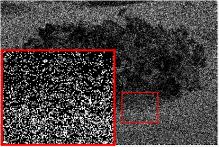}&
\hspace{-2.0ex}\includegraphics[width=0.23\linewidth]{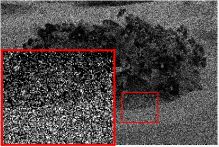} &
\hspace{-2.0ex} \includegraphics[width=0.23\linewidth]{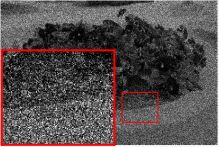} &
\hspace{-2.0ex} \includegraphics[width=0.23\linewidth]{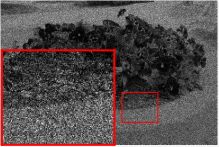}\\
\includegraphics[width=0.23\linewidth]{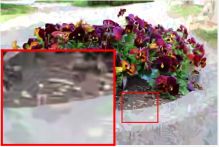}&
\hspace{-2.0ex}\includegraphics[width=0.23\linewidth]{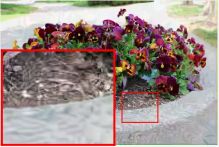} &
\hspace{-2.0ex} \includegraphics[width=0.23\linewidth]{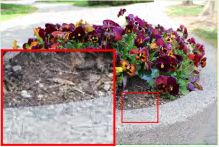} &
\hspace{-2.0ex} \includegraphics[width=0.23\linewidth]{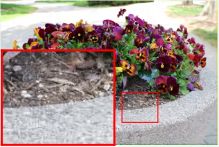}\\
\footnotesize 1-bit, $22.67$dB &
\footnotesize 2-bit, $23.86$dB &
\footnotesize 3-bit, $25.37$dB &
\footnotesize 4-bit, $26.73$dB \\
\end{tabular}
%IMG_0415_n_1_rho_1_lambda_0p0020_24p95dB_0p7659.jpg
%IMG_0415_n_2_rho_1_lambda_0p0010_25p34dB_0p8160.jpg
%IMG_0415_n_3_rho_1_lambda_0p0006_27p45dB_0p8713.jpg
%IMG_0415_n_4_rho_1_lambda_0p0004_28p18dB_0p9054.jpg

\caption{Synthetic experiment for quantitative evaluation.  [Top row]: One frame of the QIS measurements using different number of bits. [Bottom row]: Reconstructed images using the proposed method with 20 frames of QIS data.  The average photon counts per pixel are 0.25, 0.75, 1.75 and 3.75 for 1-bit, 2-bit, 3-bit and 4-bit QIS, respectively.}
\label{fig:recSingleMultiSynth}
\vspace{-1.0ex}
\end{figure*}

\subsection{Real QIS data}

We first show the reconstruction of an image of the Digital SG $\mathrm{ColorChecker}^{\footnotesize\textregistered}$ chart. We generate two sets of measurements: (a) a set of 50 one-bit frames, quantized with a threshold $q=4$, and (b) a set of 10 five-bit frames. We use PnP  and BM3D with $(\lambda,\rho)=(0.15,10)$ and $(0.01,10)$ for the 1-bit and 5-bit data, respectively. After reconstruction, the results are multiplied by a $3\times 3$ color correction matrix to mitigate any color cross-talk. This matrix is generated by linear least square regression of the 24 Macbeth color patches that lies inside the SG ColorChecker chart. The results in Fig. \ref{fig:recSingleMulti} suggest that while 1-bit mode has color discrepancy with the ground truth, the 5-bit mode is producing a reasonably-high color accuracy.

\begin{figure*}[!]
\centering
\begin{tabular}{cccc}
\includegraphics[width=0.23\linewidth]{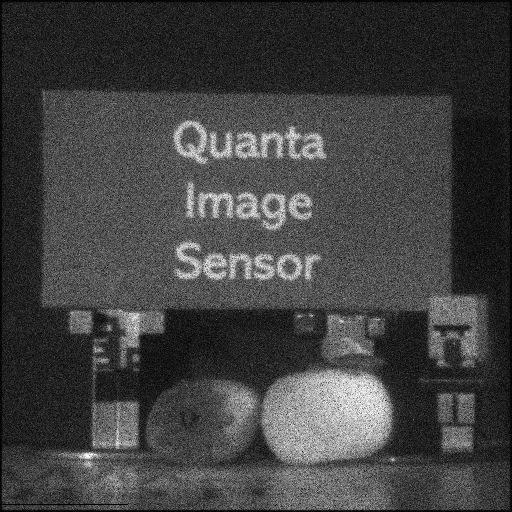}&
\hspace{-2.0ex}\includegraphics[width=0.23\linewidth]{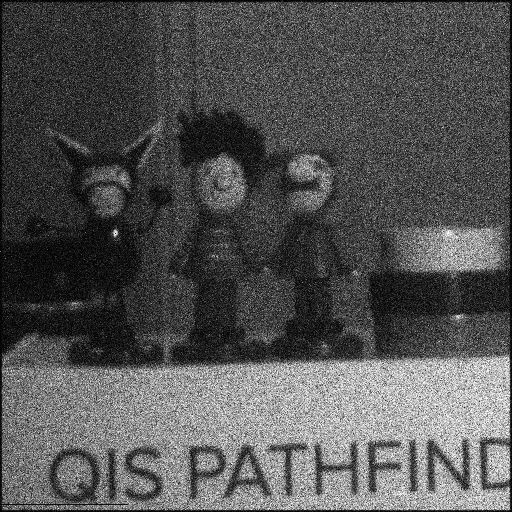} &
\hspace{-2.0ex}\includegraphics[width=0.23\linewidth]{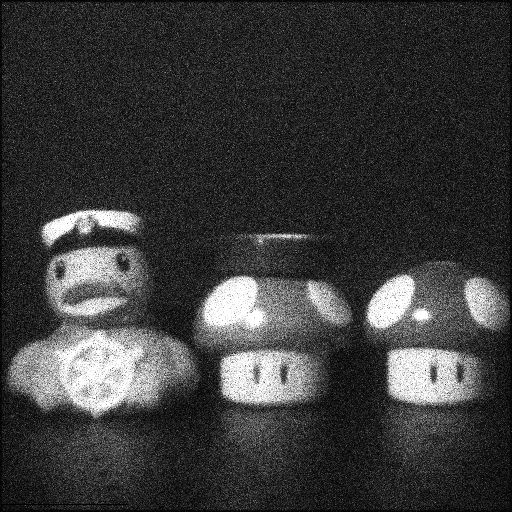}&
\hspace{-2.0ex}\includegraphics[width=0.23\linewidth]{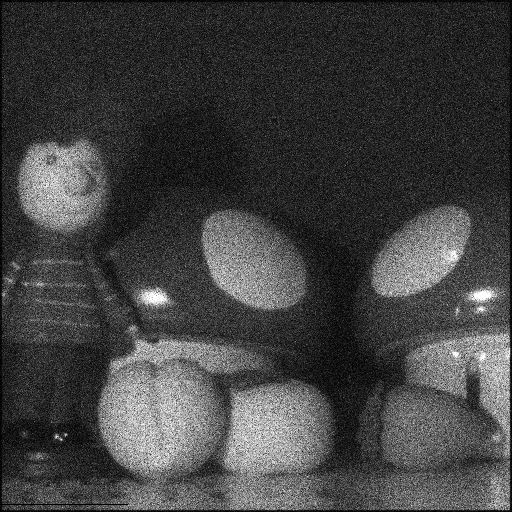}\\
\multicolumn{4}{c}{(a) Raw QIS 5-bit Data. 4 frames with exposure time of 50 $\mu$s were obtained for each scene.}\\
\includegraphics[width=0.23\linewidth]{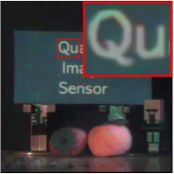}&
\hspace{-2.0ex}\includegraphics[width=0.23\linewidth]{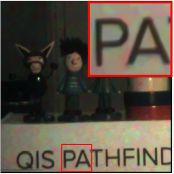} &
\hspace{-2.0ex}\includegraphics[width=0.23\linewidth]{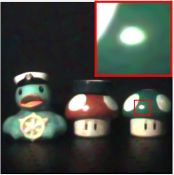} &
\hspace{-2.0ex}\includegraphics[width=0.23\linewidth]{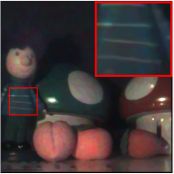}\\
\multicolumn{4}{c}{(b) Reconstruction using the proposed  method (iterative) $1024\times1024$}\\
\includegraphics[width=0.23\linewidth]{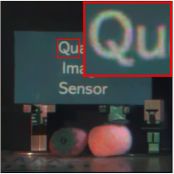}&
\hspace{-2.0ex}\includegraphics[width=0.23\linewidth]{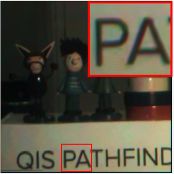} &
\hspace{-2.0ex}\includegraphics[width=0.23\linewidth]{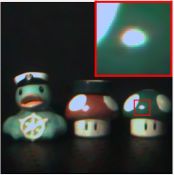} &
\hspace{-2.0ex}\includegraphics[width=0.23\linewidth]{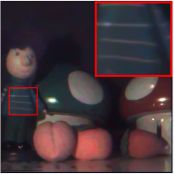}\\
\multicolumn{4}{c}{(c) Reconstruction using the proposed method (non-iterative) $512 \times 512$}
\end{tabular}
\vspace{-2ex}
\caption{ Real QIS image reconstruction. The exposure time for each frame is 50 $\mu$s. The average number of photons per frame is 4.2, 3.0, 1.9, and 2.9 for each image respectively. Both methods use 4 frames for reconstruction. The raw data has a resolution of $1024 \times 1024$ pixels. The ADMM method retains the resolution, whereas the non-iterative method reduces the resolution to $512 \times 512$. Reconstruction using both the methods are shown at the same size for easier visual comparison. Notice that the non-iterative algorithm is able to achieve a visual quality almost similar to the ADMM method.    }
\label{fig:results_reconstruction1}
\vspace{-4ex}
\end{figure*}

Next, we show the result of imaging real scenes. See   Fig. \ref{fig:results_reconstruction1}. In this experiment, the exposure time for each frame is set to 50$\mu$s. The average number of photons per pixel is approximately 4.2 photons for the ''QIS'' sign image, 3 photons for the Pathfinder image, 1.9 photons for the duck image, and 2.9 photons for the mushroom image. For all images, we collect the data using a 5-bit QIS.

We demonstrate two algorithms: (i) the proposed transform-denoise framework using PnP + BM3D, and (ii) assuming four-jots to one-pixel scenario by trading half of the spatial resolution. The typical runtime on an unoptimized MATLAB code is approximately 4 minutes for PnP, and 10 seconds for the four-jot to one-pixel method. An interesting observation is that even in the lower-resolution case, the details are not significantly deteriorated unless we zoom-in. However, the speed up we get is substantial.

\section{Conclusion and discussion}
We demonstrated the first color imaging for a mega-pixel Quanta Image Sensor (QIS). We designed a joint reconstruction-demosaicing algorithm for the quantized Poisson (and Poisson) statistics of the sensor data. Our solution involves a variance stabilizing transform and an iterative (and a non-iterative) algorithm. By integrating the sensor and the new algorithm, the sensor is able to acquire color images at a photon level as low as a few photons per pixel, a level that would be difficult for standard CMOS image sensors.

\section{Acknowledgement}
 The authors thank Prof. Eric Fossum for many valuable feedback. The 1Mjot color Quanta Image Sensor used in the demonstration was designed by J. Ma, S. Masoodian, and D. Starkey at Dartmouth College Advanced Image Sensor Laboratory led by Dr. Eric Fossum. The sensor was manufactured by Taiwan Semiconductor Manufacturing Company (TSMC). The camera module developed at Gigajot Technology was partially supported by National Science Foundation SBIR program. A Gnanasambandam, O. Elgendy and S. H. Chan are supported, in part, by the U.S. National Science Foundation under Grant CCF-1718007.

%One limitation of the current design is that we did not model color cross-talk. This explains why the reconstruction of the real image is not as satisfactory as the simulated data. Cross-talk is a phenomenon in which a photon incident on a particular pixel incites a response from a neighboring pixel. This causes degradation of the image sharpness and makes the colors fade away. One possible way to model the cross-talk is to introduce a blur matrix $\mH$ so that the observed image becomes $\vtheta = \alpha\mH\mS\mG\vc$ \cite{Anzagira_Fossum_2015}. However, since $\mH$ is not necessarily diagonal, the nonlinear Anscombe transform will cause $\calT(\mH\mS\vz) \not= \mH\mS\calT(\vz)$. As a result, there is no simple solution in the VST domain to solve the problem efficiently. Our current approach is to introduce a color correction map (a common practice in the camera industry). However, a fundamental solution would be to integrate $\mH$ into the reconstruction algorithm.

%Another approach that can potentially alleviate the color cross-talk is to design a color filter array better than the Bayer pattern. However, because of the truncated Poisson statistics and the severe cross-talk, color filter array for QIS can be very different from that of CMOS. Some prior work by Anzagira and Fossum \cite{Anzagira_Fossum_2015} has proposed various designs, yet a systematic design framework will be an interesting problem to pursue.

%\bibliographystyle{osajnl.bst}
\bibliography{egbib}
\end{document}